\documentclass{article}

\PassOptionsToPackage{numbers, compress}{natbib}

 \usepackage{graphicx}
 \usepackage{caption}
 \usepackage{subcaption}
 \usepackage{siunitx} 
 \usepackage{multirow} 


    \usepackage[preprint]{neurips_2024}

\usepackage[utf8]{inputenc} 
\usepackage[T1]{fontenc}    
\usepackage{hyperref}       
\usepackage{url}            
\usepackage{booktabs}       
\usepackage{amsfonts}       
\usepackage{nicefrac}       
\usepackage{microtype}      
\usepackage{xcolor}         

\title{Evaluating Federated Kolmogorov-Arnold Networks on Non-IID Data}

\author{%
  Arthur M. Sasse\\
  Universidade Federal do Rio de Janeiro \\
  \texttt{artsasse@cos.ufrj.br} \\
  \And
  Claudio M. de Farias \\
  Universidade Federal do Rio de Janeiro \\
  \texttt{cmicelifarias@cos.ufrj.br} \\
}

\begin{document}

\maketitle

\begin{abstract}
  Federated Kolmogorov-Arnold Networks (F-KANs) have already been proposed, but their assessment is at an initial stage. We present a comparison between KANs (using B-splines and Radial Basis Functions as activation functions) and Multi-Layer Perceptrons (MLPs) with a similar number of parameters for 100 rounds of federated learning in the MNIST classification task using non-IID partitions with 100 clients. After 15 trials for each model, we show that the best accuracies achieved by MLPs can be achieved by Spline-KANs in half of the time (in rounds), with just a moderate increase in computing time.
\end{abstract}

\section{Introduction}

Since the paper by \cite{pmlr-v54-mcmahan17a}, many works have investigated Federated Learning potential as a privacy-focused alternative to centralized Machine Learning (ML) models, offering the possibility of collaborative ML without sharing users' data, just models. Seven years later, there are still challenges to the real-life application of federated learning, like guaranteeing high accuracies across clients with non-iid data and reducing communication costs between clients and servers as described by \cite{flsurvey2024}.

In 2024, \cite{liu2024kan} proposed a new kind of neural network that could help tackle these problems. The Kolmogorov-Arnold Networks (KANs) promise higher accuracy with fewer parameters than Multi-Layer Perceptrons, the traditional neural networks. Having a completely different structure, inverting the roles of edges and nodes in a neural network, and built with learnable activation functions, KANs also offer the possibility of creating more interpretable and lean models \cite{liu2024kan}, which in turn could help scientists discover even symbolic formulas for their tasks. The paper by \cite{liu2024kan} has not yet been officially published or peer-reviewed, but since it's been made available on Arxiv\footnote{https://arxiv.org/abs/2408.10205}, about 199 studies have been produced exploring the possibilities opened by this new paradigm in ML, according to Google Scholar.

KANs have already been combined with other relevant concepts in ML, like convolutional models by \cite{convkan}, or graph neural networks by \cite{kiamari2024gkan}. Similarly, \cite{fkans_zeydan2024} and \cite{fed-wav-kan} initiated studies exploring KANs in federated learning. The potential for achieving higher accuracies with fewer parameters than MLPs and avoiding catastrophic forgetting, as described by \cite{liu2024kan}, could be useful for handling communication restraints and heterogeneous data across clients, respectively. However, there is no evidence to guarantee the convergence rate of F-KANs, and the few experimental evaluations publicly available do not explicitly address the non-iid problem, nor do they compare the F-KANs against federated MLPs with a similar number of parameters. Also, initial empirical evaluation of KANs against MLPs suggests that the execution times of the former could be ten times larger than those of the latter, threatening its use in practice.

This work aims to fill the gaps described above by conducting simulations with federated learning for KANs and MLPs under experimental conditions that are more similar to typical studies in FL. This study is a fundamental step in deciding whether we can integrate KANs into the current FL ecosystem, which specific adaptations they may need due to their structural differences from the classical MLPs, and if KANs can be considered valuable tools for solving FL practical problems. Our focus is evaluating whether KANs can also achieve higher accuracies than MLPs in federated learning and the relationship between both models' computing times.

First, we compare KANs and MLPs (with a similar number of parameters) concerning the global model test accuracy achieved over 100 rounds of FL in the MNIST digits classification task. The dataset is partitioned over 100 clients in a "pathological" non-iid way, with each client having data for only two labels. For example, one client has data for some of the handwritten digits "1" and "2", while another has part of the data from digits "7" and "8". For KANs, we test two kinds of activation functions: B-Splines, as initially proposed by \cite{liu2024kan}, and Radial Basis Functions (RBFs), as proposed by \cite{radialbasis}. We use FedAvg with momentum as the aggregation algorithm to achieve provable convergence for the non-iid data, as described by \cite{cheng2024momentumbenefitsnoniidfederated}.

After 100 rounds of communication, we show, with statistical significance, that Spline-KANs achieve higher accuracies than MLPs during all phases of federated learning, particularly during the first half of the simulation. RBF-KANs perform poorly compared to the other models during most rounds, suggesting we may need to adjust the aggregation algorithm according to each KAN activation function. The execution times from the KAN models are indeed higher than the MLPs' times, as expected, but much lower than initially reported in earlier studies, probably due to the new and more efficient implementations.

The document is divided as follows: in Section \ref{related}, we compare this work with others in the field of federated KANs; in Section \ref{methodology}, we describe our experimental setup and statistical tests; in Section \ref{results} we present graphs and the results of our tests, discussing our findings; in Section \ref{limitations} we comment on the limitations of our study; and in Section \ref{conclusion} we synthesize our findings and perspectives about federated KANs.

\section{Related Works}
\label{related}

The original KAN paper by \cite{liu2024kan} presented the theory behind these networks and focused on empirical evaluations with physical and mathematical problems having small associated datasets. Our study focuses on a federated setup with a much larger dataset (MNIST), which is classically used for model evaluation in the Machine Learning area. 

\cite{radialbasis} presented a new type of KAN, using radial basis functions as activations functions, and made the implementation available in GitHub as a package for Python called FastKAN. The author claims that FastKAN is faster than the traditional B-Spline implementation based on an evaluation with a centralized MNIST classification task. We use FastKAN to classify our pathologically partitioned MNIST dataset.

\cite{fkans_zeydan2024} coined the term "F-KAN", presenting the method for federating KANs, based on the FedAvg algorithm for MLPs. The authors evaluated the proposed model through a classification task using the Iris dataset \cite{iris_53}. The Iris dataset consists of data for 50 samples of flowers from three different species, described by four features. The dataset is partitioned between two clients in an iid way. The simulation comprised one execution for 20 rounds of federated learning, comparing an MLP and a Spline-KAN model, each with three hidden layers of 20 neurons. The KAN model showed superior performance for four metrics: accuracy, recall, precision, and F1-score. This work also compares KANs and MLPs in a federated setting. However, we change some parameters in trying to simulate a more realistic FL configuration: instead of the Iris Dataset, we use MNIST, which comprises 70,000 examples (60,000 for training and 10,000 for testing) from 10 classes of handwritten digits, represented by images with 28*28 pixels (784 features); we divide the data over 100 clients in a non-iid manner, and only 10\% of clients participate in each round, simulating clients that are not always available; we run 15 executions for each model to achieve more statistically significant results; we compare MLPs not only against Spline-KANs but also against RBF-KANs; we run 100 communication rounds to capture more of each model behavior through time; and we reduce the number of KAN neurons so both models have a similar number of parameters. Table \ref{tab:comparison} shows the detailed comparison between each study.

\begin{table}[h]
\caption{Comparison of experimental setups between \cite{fkans_zeydan2024} and our work.}
\label{tab:comparison}
\centering
\begin{tabular}{lcc}
\toprule
\textbf{Parameter} & \textbf{\cite{fkans_zeydan2024}} & \textbf{Our Work} \\ 
\midrule

\multicolumn{3}{c}{\textbf{General}} \\
\midrule
Dataset & Iris & MNIST \\
Classes & 3 & 10 \\
Features & 4 & 28 * 28 \\
Samples & 150 & 60,000 + 10,000 \\
Clients & 2 & 100 \\ 
Partition & iid & non-iid \\
Rounds & 20 & 100 \\ 
Batch size & 16 & 64 \\ 
Local Epochs & 20 & 5 \\
Learning rate & 0.001 & 0.1 \\ 
Clients per round  & 100\% & 10\% \\
Weight decay & $1 \times 10^{-5}$ & -- \\ 
Dropout probability & 0.5 & -- \\ 
Aggregation & FedAvg & FedAvg \\
Optimizer & Adam & SGD with momentum = 0.9 \\
KAN Activation & B-Spline & B-Spline, RBF \\
Executions per model & 1 & 15 \\

\midrule
\multicolumn{3}{c}{\textbf{KAN}} \\
\midrule
Total Parameters (Spline) & 12,383 & 196,320 \\
Total Parameters (RBF) & -- & 178,410 \\
Hidden layer sizes & [20, 20, 20] & [24, 24] \\ 
Grid size & 5 & 5 \\ 
Spline order & 3 & 3 \\ 
RBF Centers & -- & 8 \\

\midrule
\multicolumn{3}{c}{\textbf{MLP}} \\
\midrule
Total parameters & 1,003 & 199,210 \\ 
Hidden layer sizes & [20, 20, 20] & [200, 200]\\ 

\bottomrule
\end{tabular}
\end{table}

\cite{fed-wav-kan} present a study in progress about federated learning with KANs, which use different kinds of wavelets as their activation functions, based on \cite{Bozorgasl_2024}. The authors show partial results from a federated Mexican hat Wavelet-KAN in a [28*28, 64, 10] layer arrangement for the MNIST dataset classification. They claim Wavelet-KANs are superior to Spline-KANs. However, direct comparisons between models were not available at the time of this writing. The results are averaged over a non-disclosed number of trials for each number of clients (10, 20, and 50) and also for a centralized model. The authors specify that each model is trained over 50 epochs, but we could not find how many communication rounds happened in each simulation. Besides differences in the experimental setups, our paper is focused only on Spline-KANs and RBF-KANs, not including Wavelet-KANs. The Wavelet-KAN implementation is available in GitHub but not as a Python package, which created some difficulties adapting it to our study, which we hope to overcome in future studies.

Our work differentiates itself by presenting a broader set of experiments on federated KANs, drawing our conclusions from more common settings in the ML and FL areas. We believe this work can serve as a baseline for future experiments comparing KANs and MLPs' capacities in federated architectures.

\section{Methodology}
\label{methodology}

This work is guided by two research questions and their respective metrics and hypotheses.

\begin{itemize}
    \item \textbf{Research Question 1}
    \begin{itemize}
        \item \textbf{Question}: Do KANs achieve higher test accuracies than MLPs across 100 rounds of federated learning in the MNIST classification task?
        \item \textbf{Metric}: Test accuracy of the global models for rounds [10, 20, 30, 40, 50, 60, 70, 80, 90, 100].
        \item \textbf{Null Hypothesis}: KANs achieve a test accuracy lower than or equal to that of MLPs in a given round of federated learning.
        \item \textbf{Alternative Hypothesis}: KANs achieve a higher test accuracy than MLPs in a given round of federated learning.
        \item \textbf{Hypothesis Test}: Welch's t-test \cite{welch1947}, since we can't assume the models' accuracies will have the same variances, and it's a robust test against deviations from normality \cite{welch}.
    \end{itemize}
    \item \textbf{Research Question 2}
    \begin{itemize}
        \item \textbf{Question}: What is the relationship between the execution times of KANs and MLPs for 100 rounds of federated learning in the MNIST task?
        \item \textbf{Metric}: Execution time, in seconds.
        \item \textbf{Method}: Calculate the confidence interval for the ratio between the execution times of KANs and MLPs, using bootstrapping \cite{bootstrap} (10,000 bootstrapped samples of size 15).
    \end{itemize}
\end{itemize}

\subsection{Experimental Setup}

To make a fair comparison between KANs and MLPs, we defined a maximum number of parameters for each model. First, we fix the architecture of the MLP classifier, inspired by one of the models studied on the FedAvg paper by \cite{pmlr-v54-mcmahan17a}, a [28*28, 200, 200, 10] neural network. Based on that, we built a Spline-KAN model with the same number of layers, calculating the number of neurons needed in each layer to achieve, at most, the same number of parameters used by the MLP, resulting in a [28*28, 24, 24, 10] network. We repeated this architecture for the RBF-KAN. Table \ref{tab:comparison} details the resulting architectures and hyperparameters.

We implemented our simulation using PyTorch (\cite{Ansel_PyTorch_2_Faster_2024}) and the Flower (\cite{beutel2022flowerfriendlyfederatedlearning}) framework for federated learning. We used the efficient-kan implementation (\cite{blealtan_efficient_kan}) of Spline-KANs, whose contribution to enhancing KAN's performance has already been acknowledged by the authors of the original KAN paper in a subsequent article (\cite{liu2024kan2}). For RBF-KANs, we used \cite{radialbasis} implementation. 
Our code repository is available at \url{https://github.com/artsasse/fedkan}.

All simulations were run in the Google Colaboratory \footnote{https://colab.google} environment, with an Intel Xeon CPU @ 2.20 GHz with two virtual CPUs and 13 GB RAM, alongside a NVIDIA Tesla T4 GPU with 15GB VRAM.

The local models were aggregated by the server using the FedAvg algorithm (\cite{pmlr-v54-mcmahan17a}) with momentum, since \cite{cheng2024momentumbenefitsnoniidfederated} proved how using momentum can make FedAvg converge without needing to use decaying learning rates, thus simplifying our setup.

\begin{figure}[h]
  \centering
  \includegraphics[scale=0.4]{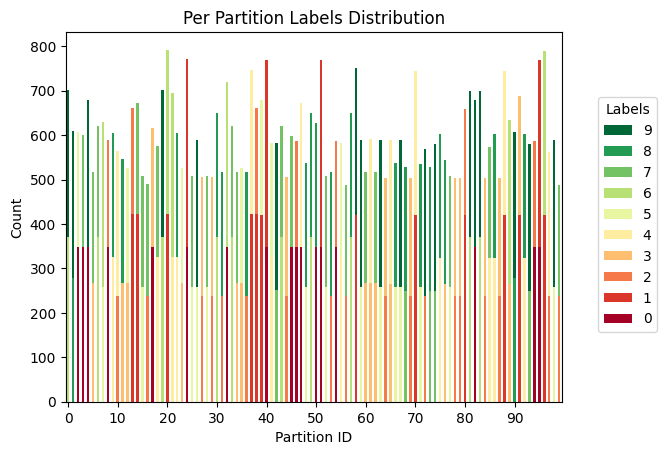}
  \caption{Per partition labels distribution.}
  \label{fig:partitions}
\end{figure}

The MNIST (\cite{mnist}) is a handwritten digits dataset commonly used for evaluating classification models. We used its original split, with 60,000 samples for training and 10,000 for testing. We kept the test samples together for centralized evaluation of the global model at the end of each round of FL. The training set was "pathologically" partitioned between 100 clients so that each client holds examples for only two digits, guaranteeing the non-iid property, a common challenge for real-life FL applications. Figure \ref{fig:partitions} shows graphically how the data was partitioned. Overall, the partitions are slightly balanced: the number of examples for each client varies between 488 and 792, with mean = 600. This skewed division was inspired by \cite{pmlr-v54-mcmahan17a} and \cite{li2020federatedoptimizationheterogeneousnetworks}. However, the former used balanced partitions, and the latter created unbalanced partitions following a power law.

Since we only select a fraction of the clients for training in each round of FL, the training accuracy is calculated using different data for each round of communication. Thus, its values could change dramatically from one round to the next even if the global model stayed the same.

\section{Results}
\label{results}

\begin{figure}[h]
  \centering
  \includegraphics[scale=0.4]{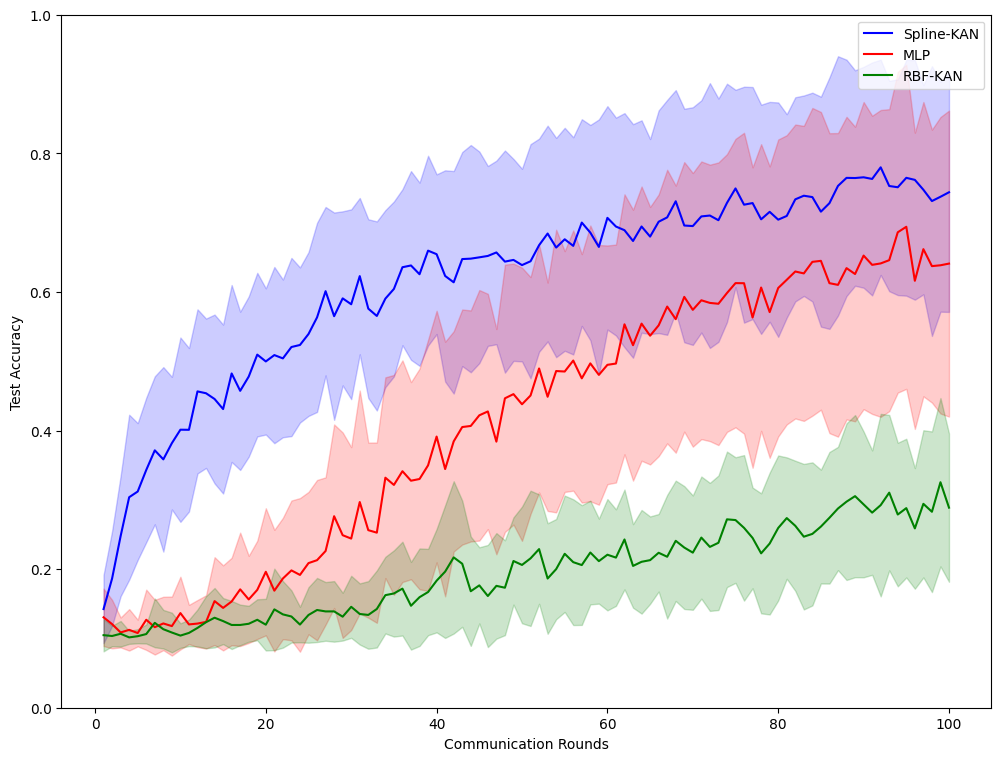}
  \caption{Mean test accuracy for 100 rounds, with error bands.}
  \label{fig:results}
\end{figure}

\begin{figure}[h]
  \centering
  \begin{subfigure}[b]{0.32\textwidth}
    \includegraphics[width=\textwidth]{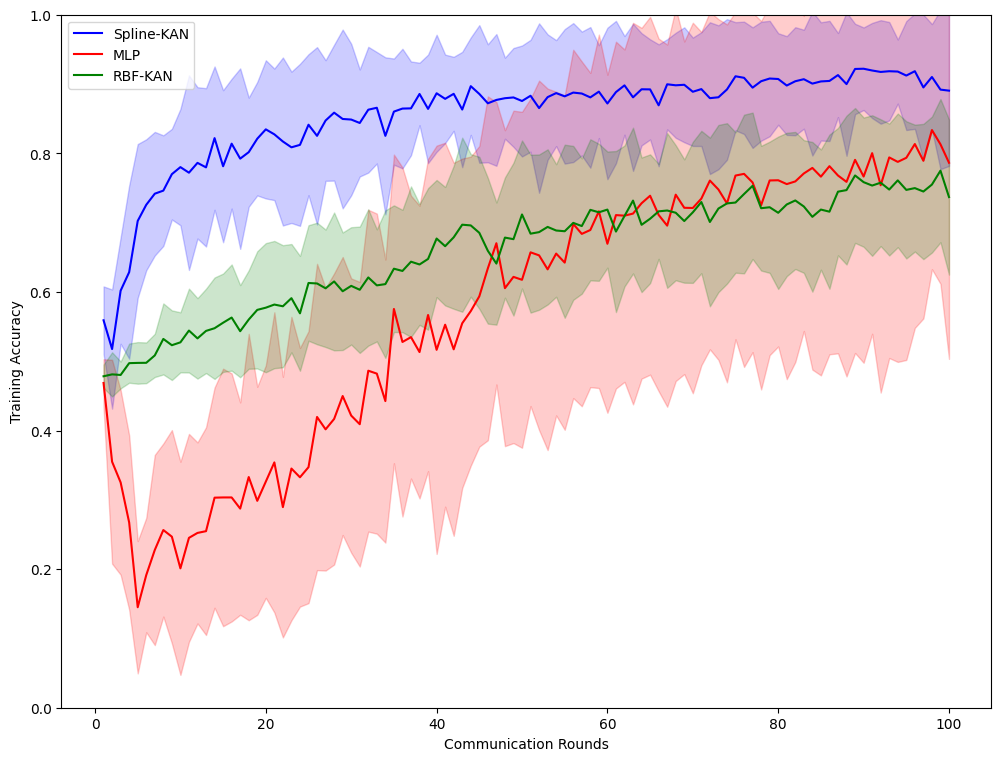}
    \caption{Train Accuracy}
    \label{fig:sub1}
  \end{subfigure}
  \hfill
  \begin{subfigure}[b]{0.32\textwidth}
    \includegraphics[width=\textwidth]{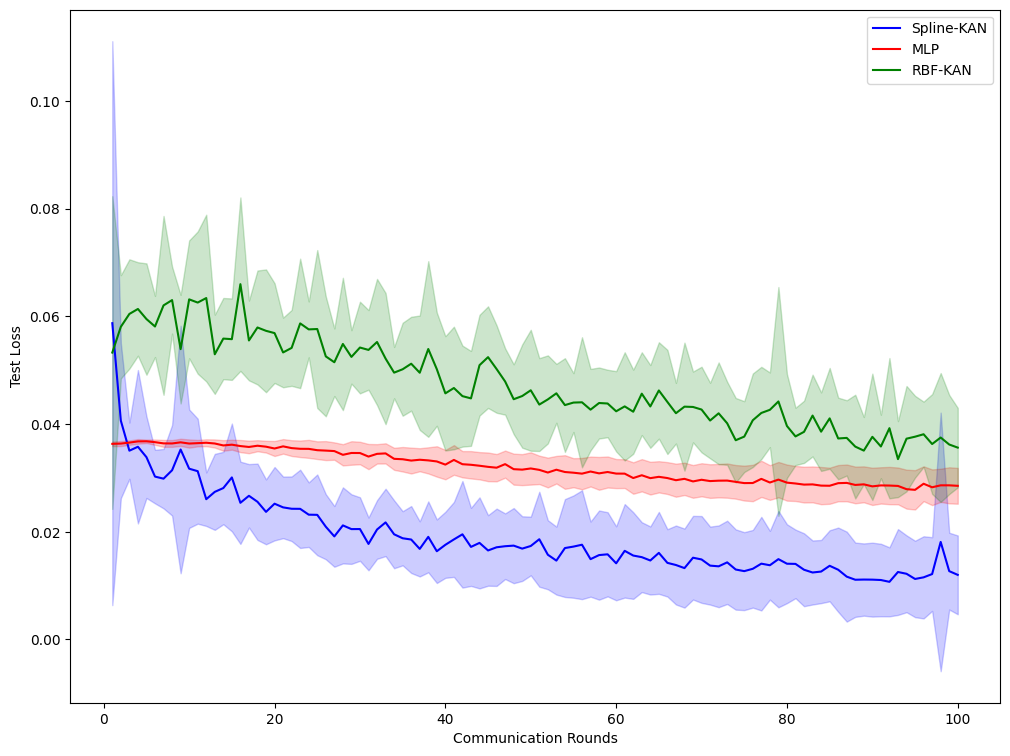}
    \caption{Test Loss}
    \label{fig:sub2}
  \end{subfigure}
  \hfill
  \begin{subfigure}[b]{0.32\textwidth}
    \includegraphics[width=\textwidth]{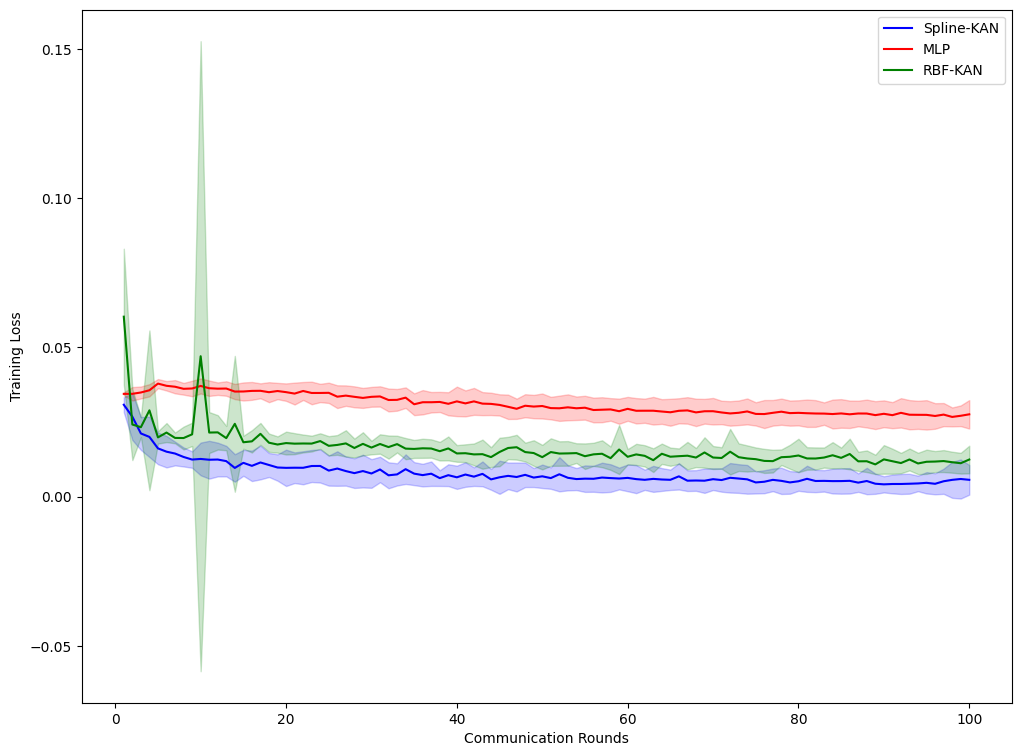}
    \caption{Train Loss}
    \label{fig:sub3}
  \end{subfigure}
  \caption{Mean accuracy and loss metrics for 100 rounds, with error bands.}
  \label{fig:four_images}
\end{figure}

To answer our \textbf{Research Question 1}, we gathered the test accuracy metric, represented in the graph from Figure \ref{fig:results}, for each of the 100 rounds of FL. Intuitively, the Spline-KAN appears to have the overall best performance, followed by the MLP, while the RBF-KAN shows very low accuracy, even in the final rounds of communication. For completeness, in Figure \ref{fig:four_images}, we also represent the training accuracy (\ref{fig:sub1}), the test loss (\ref{fig:sub2}), and the training loss (\ref{fig:sub3}), which show tendencies aligned with what we observe for the test accuracy.

Since the intersecting error bands can confound our visual analysis, we also present the test accuracies for ten different rounds of our simulation in Table \ref{tab:test_accuracy}, showing each model's mean and standard error. The results of the actual hypotheses tests are represented further in Table \ref{tab:p_values}, where we present the p-values for the Welch's t-test \cite{welch1947}. Comparing Spline-KANs and MLPs, for a 90\% significance level, we can refute the null hypothesis for all the selected rounds. For a 95\% significance level, we refute the null hypothesis from round 10 to 70. The differences between mean test accuracies are greater than 20\% from round 10 to round 60. Besides the statistically significant superiority of Spline-KANs during all the selected rounds, we note that the Spline-KAN's test accuracy increases faster during the first half of the simulation, slowing down to almost a linear increase for the second half. The MLPs presented a practically linear increase during the 100 rounds. On average, the best accuracies are achieved by the MLPs during the last two rounds, while similar values had already been achieved by Spline-KANs in round 40. The RBF-KANs also showed a linear-like increase in its metric, but much slower than the MLPs, reaching a maximum average accuracy a little higher than 0.2. We conjecture that RBF-KANs may need other aggregation algorithms, more adapted to its parameters than FedAvg, to reach good accuracies, which should be investigated in future works.

\begin{table}[h]
    \centering
    \caption{Test accuracy over communication rounds.}
    \label{tab:test_accuracy}
    \begin{tabular}{
        c
        S[table-format=1.3]@{\,\( \pm \)\,}S[table-format=1.3]
        S[table-format=1.3]@{\,\( \pm \)\,}S[table-format=1.3]
        S[table-format=1.3]@{\,\( \pm \)\,}S[table-format=1.3]
    }
        \toprule
        \multirow{2}{*}{\textbf{Round}} & \multicolumn{2}{c}{\textbf{Spline-KAN}} & \multicolumn{2}{c}{\textbf{RBF-KAN}} & \multicolumn{2}{c}{\textbf{MLP}} \\
        & \textbf{Mean} & \textbf{Std Error} & \textbf{Mean} & \textbf{Std Error} & \textbf{Mean} & \textbf{Std Error} \\
        \midrule
        10  & 0.401 & 0.138 & 0.104 & 0.018 & 0.137 & 0.054 \\
        20  & 0.500 & 0.109 & 0.120 & 0.039 & 0.196 & 0.095 \\
        30  & 0.582 & 0.142 & 0.146 & 0.046 & 0.244 & 0.137 \\
        40  & 0.654 & 0.119 & 0.183 & 0.077 & 0.391 & 0.188 \\
        50  & 0.639 & 0.144 & 0.206 & 0.087 & 0.438 & 0.204 \\
        60  & 0.707 & 0.167 & 0.221 & 0.083 & 0.495 & 0.178 \\
        70  & 0.695 & 0.177 & 0.224 & 0.085 & 0.574 & 0.204 \\
        80  & 0.704 & 0.175 & 0.259 & 0.108 & 0.606 & 0.222 \\
        90  & 0.765 & 0.165 & 0.293 & 0.109 & 0.652 & 0.229 \\
        100 & 0.744 & 0.179 & 0.289 & 0.111 & 0.641 & 0.229 \\
        \bottomrule
    \end{tabular}
\end{table}

\begin{table}[h]
    \centering
    \caption{Mean accuracy differences and p-values over communication rounds for Spline-KAN compared to MLP.}
    \label{tab:p_values}
    \begin{tabular}{
        c
        S[table-format=1.3]
        S[table-format=1.3]
    }
        \toprule
        \textbf{Round} & \textbf{Acc$_{\text{Spline}}$ $-$ Acc$_{\text{MLP}}$} & \textbf{Acc$_{\text{Spline}}$ $>$ Acc$_{\text{MLP}}$ (p-value)} \\
        \midrule
        10  & 0.265 & 0.0000 \\
        20  & 0.304 & 0.0000 \\
        30  & 0.338 & 0.0000 \\
        40  & 0.263 & 0.0001 \\
        50  & 0.201 & 0.0023          \\
        60  & 0.212 & 0.0011          \\
        70  & 0.121 & 0.0474          \\
        80  & 0.099 & 0.0940          \\
        90  & 0.113 & 0.0669          \\
        100 & 0.103 & 0.0905          \\
        \bottomrule
    \end{tabular}
\end{table}

First of all, our results are positive evidence for using Spline-KANs in FL, since they outperformed, during all the simulation phases, the MLPs, which are the original target of FedAvg and the other FL tools used in this work. Also, the accelerated increase in accuracy in early rounds could be an interesting feature for federated systems that do not prioritize a very high accuracy but aim to minimize the training rounds needed to reach a good enough generalization. This limitation could be due to communication or battery restraints, common for sensor networks and the Internet of Things \cite{flsurvey2024}. The model compressibility and energy usage are properties that should be further studied when considering the application of KANs in these domains \cite{flsurvey2024}.

Regarding \textbf{Research Question 2}, we can observe the analysis results in Table \ref{tab:execution_time}. We can see that RBF-KANs have execution times similar to those of MLPs. And, even though Spline-KANs have greater variance in their execution times and can be almost 1.36$\times$ slower than MLPs, that difference is not as dramatic as expected. In \cite{liu2024kan} and \cite{fkans_zeydan2024}, the authors related that KANs had taken 10x or even 100x longer than MLPs to execute. We note that both works reported results from the initial implementation of KANs. In our work, we use implementations of KAN, \cite{blealtan_efficient_kan} for Spline-KAN, and \cite{radialbasis} for RBF-KAN, built specially to tackle these performance issues. Moreover, let's consider cases where we want the global model to achieve a specific accuracy. The Spline-KANs may reach that target much earlier than MLPs regarding communication rounds, which would translate to an earlier stoppage in execution and a shorter computing time. In our case, if our goal were a 65\% accuracy, MLPs would have stopped by round 90 and Spline-KANs by round 40, on average.

\begin{table}[ht]
    \centering
    \caption{Execution time analysis for different models over 100 rounds}
    \label{tab:execution_time}
    \begin{tabular}{
        l
        c
        c
        c
    }
        \toprule
        \textbf{Model} & \textbf{Mean Time (s)} & \textbf{Std Error} & \textbf{Ratio (95\% CI)} \\
        \midrule
        MLP         & 3593 & 124  & \(1.00\times\) \\
        Spline-KAN  & 4192 & 1183 & \(1.02\times\,\text{--}\,1.36\times\) \\
        RBF-KAN     & 3640 & 114  & \(0.99\times\,\text{--}\,1.04\times\) \\
        \bottomrule
    \end{tabular}
\end{table}

\section{Limitations}
\label{limitations}
A more realistic way to assess federated learning concerning data partitioning would have been to utilize the FEMNIST (Federated Extended MNIST) dataset \cite{caldas2019leaf}. Instead of creating an artificial posthoc partition, FEMNIST gives each client access to data from just one handwritten digit author, generating naturally non-i.i.d partitions. In the future, we plan to evaluate F-KANs on FEMNIST and other tasks besides digit classification.

Convolutional models are generally considered a better option for classifying images than fully connected networks \cite{huang2018denselyconnectedconvolutionalnetworks}. We are unsure if using convolutional networks would cause the disparity between MLPs and KANs to grow or shrink. Even though convolutional KANs have already been proposed \cite{convkan}, because these works are relatively recent, we believe potential flaws in these early implementations could skew the results and jeopardize the study's validity.

There is still no standard method for comparing KANs and MLPs for the same tasks. In an effort to provide a fair comparison in terms of each model's generalization capacity, we decided to utilize a similar number of parameters. On the other hand, arbitrary choices made about the construction of both models could have affected the outcomes in unknown ways. In future works, we intend to perform a hyperparameter optimization to make a fairer comparison between each type of neural network. Also, there are still other activation functions available for KANs that were not investigated in this work, like wavelets \cite{wavkan}, ReLU \cite{relukan}, Chebyshev \cite{chebyshevkan}, Jacobi \cite{jacobikan} \cite{jacobikan2} and Fourier \cite{fourierkan} functions. 

Since MLPs have been used much longer than KANs, many specific software and hardware optimizations not used in this work might drastically reduce the MLP execution times. However, if KANs become a popular model for the machine learning community, new optimizations and performance tricks are expected to be created for them, too.

\section{Conclusion}
\label{conclusion}
In this work, we compared two types of KANs against an MLP model over 100 rounds of federated learning for image classification. We focused on the test accuracy for many simulation phases and the total execution time. Spline-KANs performed better than MLPs during all the simulation phases, especially during the first 60 rounds, showcasing how these types of KANs could be an alternative to achieve higher accuracies in fewer rounds of FL. RBF-KANs reached accuracies well below those of MLPs, demonstrating how the KAN activation function may strongly impact its performance. Both KAN implementations observed execution times at most 1.36$\times$ the MLP's execution time, which is way faster than initial implementations of KAN achieved in centralized and federated simulations. Before obtaining a theoretical explanation for the differences between each model, it is impossible to estimate the generality of these results for different FL configurations. Regardless, this highly significant result indicates that KANs could be an interesting path to solve some of the most relevant problems in federated learning.

\begin{ack}
We thank Larissa Galeno and Juliana Dal Piaz for the helpful discussions and Paulo Cesar Silva, Ana Tostes, and Igor Mota da Costa for their support in running the simulations.

Arthur M. Sasse’s work is supported by a scholarship from CAPES Foundation, Brazil, from the Ministry of Education of Brazil.

This work was partially funded by FAPERJ (grant 201.467/2022).
\end{ack}

\medskip

\bibliographystyle{plainnat}
\bibliography{bibliography}


\appendix


\end{document}